\documentclass[letterpaper, 10 pt, conference]{ieeeconf}  

\IEEEoverridecommandlockouts                              

\usepackage{amsmath}
\usepackage{graphicx}
\usepackage[breaklinks=true,colorlinks=true,linkcolor=blue,urlcolor=blue,citecolor=blue]{hyperref}
\usepackage{mathptmx} 
\usepackage{amssymb}  
\usepackage{caption,subcaption}
\usepackage{mathtools}
\usepackage[]{units}
\usepackage{array}
\usepackage[top=1in, bottom=1.in, left=1.0in, right=1.0in]{geometry}
\newcommand{\squeezeup}{\vspace{-3mm}}

\title{\LARGE \bf
Body Lift and Drag for a Legged Millirobot in Compliant Beam Environment
}

\author{Can Koc$^{*1}$, Cem Koc$^{*1}$, Brian Su$^{*2}$, Carlos S. Casarez$^{3}$ and Ronald S. Fearing$^{1}$
\thanks{ This work was supported by the United States Army Research Lab under the Micro Autonomous Science and Technology Collaborative Alliance.}
\thanks{$^{1}$Can Koc, Cem Koc and Ronald S. Fearing are with the Department of Electrical Engineering and Computer Sciences, 
        {\tt\small cankoc@berkeley.edu,cemkoc@berkeley.edu,
        ronf@berkeley.edu}}%
\thanks{$^{2}$Brian Su is with the Department of Computer Science, 
        {\tt\small bsu@berkeley.edu}}%
\thanks{$^{3}$Carlos S. Casarez is with the Department of Mechanical Engineering.
 {\tt\small casarezc@berkeley.edu}
 All  University of California, Berkeley, CA 94720 USA.
       }%
\thanks{$^{*}$These authors contributed equally to this work.}%
  }

\begin{document}

\maketitle
\thispagestyle{empty}
\pagestyle{empty}

\begin{abstract}

Much current study of legged locomotion has rightly focused on foot traction forces, including on granular media. Future legged millirobots will need to go through terrain, such as brush or other vegetation, where the body contact forces significantly affect locomotion. In this work, a (previously developed) low-cost 6-axis force/torque sensing shell is used to measure the interaction forces between a hexapedal millirobot and a set of compliant beams, which act as a surrogate for a densely cluttered environment. Experiments with a VelociRoACH robotic platform are used to measure lift and drag forces on the tactile shell, where negative lift forces can increase traction, even while drag forces increase. The drag energy and specific resistance required to pass through dense terrains can be measured. Furthermore, some contact between the robot and the compliant beams can lower specific resistance of locomotion. For small, light-weight legged robots in the beam environment, the body motion depends on both leg-ground and body-beam forces. A shell-shape which reduces drag but increases negative lift, such as the half-ellipsoid used, is suggested to be advantageous for robot locomotion in this type of environment.

\end{abstract}

\section{INTRODUCTION}
Recently developed low-cost folded legged milli-robots have demonstrated high-speed locomotion~\cite{Haldane2015Regime} and turning maneuverability on diverse terrain~\cite{SwarmHaldane}~\cite{Casarez2016}, which makes them potential candidates for performing a swarm-like distributed search for survivors in disaster scenarios such as an earthquake building collapse. This class of palm-sized robots is manufactured using the scaled Smart Composite Microstructures (SCM) process, which enables inexpensive and rapid fabrication of millimeter-scale integrated folded mechanisms that drive many legs with few motors~\cite{Hoover2008}~\cite{Haldane2013}. Folded legged millirobots are lightweight and compact, which allows them to pass through tight spaces and traverse rubble with minimal disturbance to the environment.

\begin{figure}[t]
\addtolength{\belowcaptionskip}{-5mm}
\centering
\begin{subfigure}[b]{0.48\textwidth}
    \centering
    \includegraphics[width=0.7\linewidth]{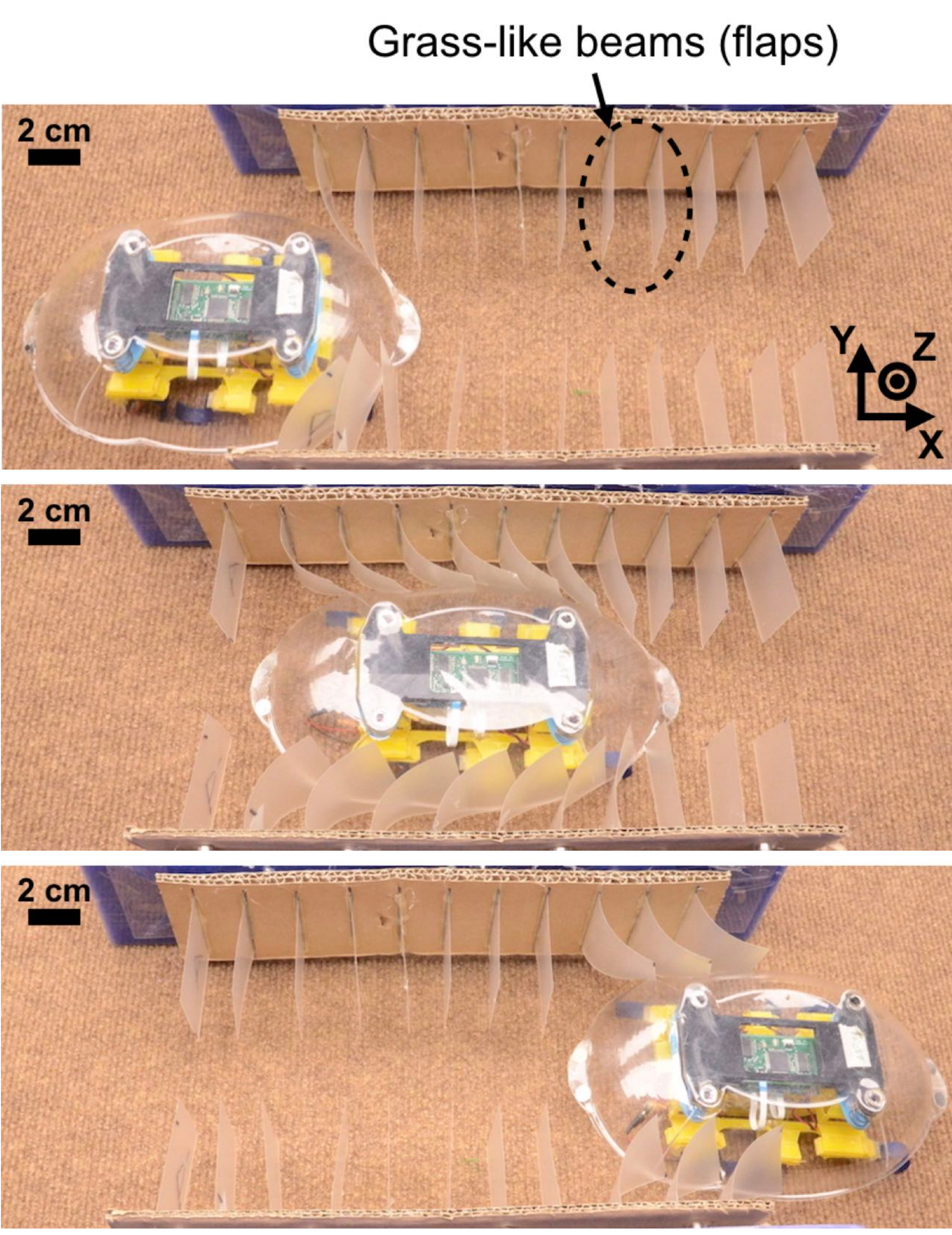}
\end{subfigure}
\caption{VelociRoACH with tactile shell traversing a channel with compliant beams spaced for lateral gap of 6 cm.}
\label{fig:grass_experiment}
\end{figure}

In heterogeneous environments, the size, shape, material composition, and density of obstacles can affect a millirobot's ability to traverse terrain. To develop optimal traversal routes, sensors such as cameras, lidars, ultrasonic rangefinders and force-torque sensors can be mounted on robots to map the environments and measure energetic cost of transport. Each of these sensors presents  tradeoffs with respect to cost, functionality, and precision. While recent advancements in computer vision have enabled high-precision object detection, low light conditions and particulate fouling of lenses in collapsed building environments can degrade camera visibility. Lidars are capable of precisely mapping the shape and distance of nearby objects but are expensive and too large to be mounted on millirobots. 

For small robots which are unlikely to damage themselves or the environment, force-torque sensors can be used to identify the shape or overall body force of contacts with obstacles. For locomotion with body contact with the surroundings, robot gait design and body/shell design are also important in developing shape-based controllers that can aid in efficient traversal of terrain~\cite{Travers2016}. Certain gaits have been found to be more effective and efficient in various heterogeneous and dynamic environments. Recent work has also found that shell shape affects the ability of small insects and millirobots to traverse through densely cluttered terrain~\cite{Chen2015}.

In this work, we use a previously developed force-torque sensor~\cite{Goldberg2015} that detects the spring-loaded deflections of a shell mounted to a hexapedal millirobot to measure the contact forces. These contact forces are exerted on the shell while traversing a channel with grass-like beams at varying clutter densities shown in Fig.~\ref{fig:grass_experiment}. We analyze contact forces collected from the robot traversing the channel to resolve terradynamic drag and lift forces that act on the robot. To the best of our knowledge, this work is the first to directly measure environmental drag, lift, and specific resistance of legged locomotion with terrain contact impeding forward progress.
\vspace{-3pt}
\section{RELATED WORK}
While tactile sensing has mainly been aimed at robot hands, there has been some research focused on tactile sensing in mobile robots. This work can be divided into the two categories ---distributed contact sensing and net force-torque sensing. Distributed contact sensing has been implemented using artificial antennae, whiskers and hair sensors~\cite{Cowan2005},~\cite{Jung1996},~\cite{Karras2012}. Force-torque sensing has been achieved utilizing rigid outer shells attached to the body of the robot~\cite{Goldberg2015},~\cite{Tsuji2008},~\cite{Tsuji2009} where rigid body mechanics were employed to achieve force-torque measurements~\cite{Bicchi1993}. While both modes have their uses, here we focus on understanding the effect of net loads rather than discriminating geometry features, and hence employ shell force-torque sensing, as introduced in Goldberg et al.~\cite{Goldberg2015}. \par
Previous work on locomotion through grass-like terrains focuses on characterizing how body shape affects terrain traversal~\cite{Chen2015},~\cite{2016Tanaka}. Li et al.~\cite{Chen2015} used different shell shapes in legged robot and insect locomotion experiments to reach the conclusion that rounded, more terradynamically streamlined shapes increased the chance of traversal of densely cluttered vertical grass-like beams. Gart et al.~\cite{Gart2018},~\cite{Gart2018-2} employed characteristics of gaps and bumps to model traversal patterns of a legged millirobot. However, these previous studies did not focus on measuring environmental drag energy or forces in locomotion. \par
Cost of Transport (CoT) is an important metric for mobile robot duration and range. Previous work in understanding the cost of transport for legged robots considers mainly body and leg dynamics, foot contact loading, leg trajectories, leg compliance, and actuation efficiency. For example Zarrouk et al. ~\cite{Zarrouk2013} considered the cost of transport for a milli-robot with in-plane compliant leg motion. This work expands the study of CoT to cluttered environments with significant drag forces generated by contact between a robot's body and the environment. 

\section{FORCE SENSING OVERVIEW}
\begin{figure}
\addtolength{\belowcaptionskip}{-5mm}
    \begin{tabular}{cc}
        \centering
        \includegraphics[width=0.47\linewidth]{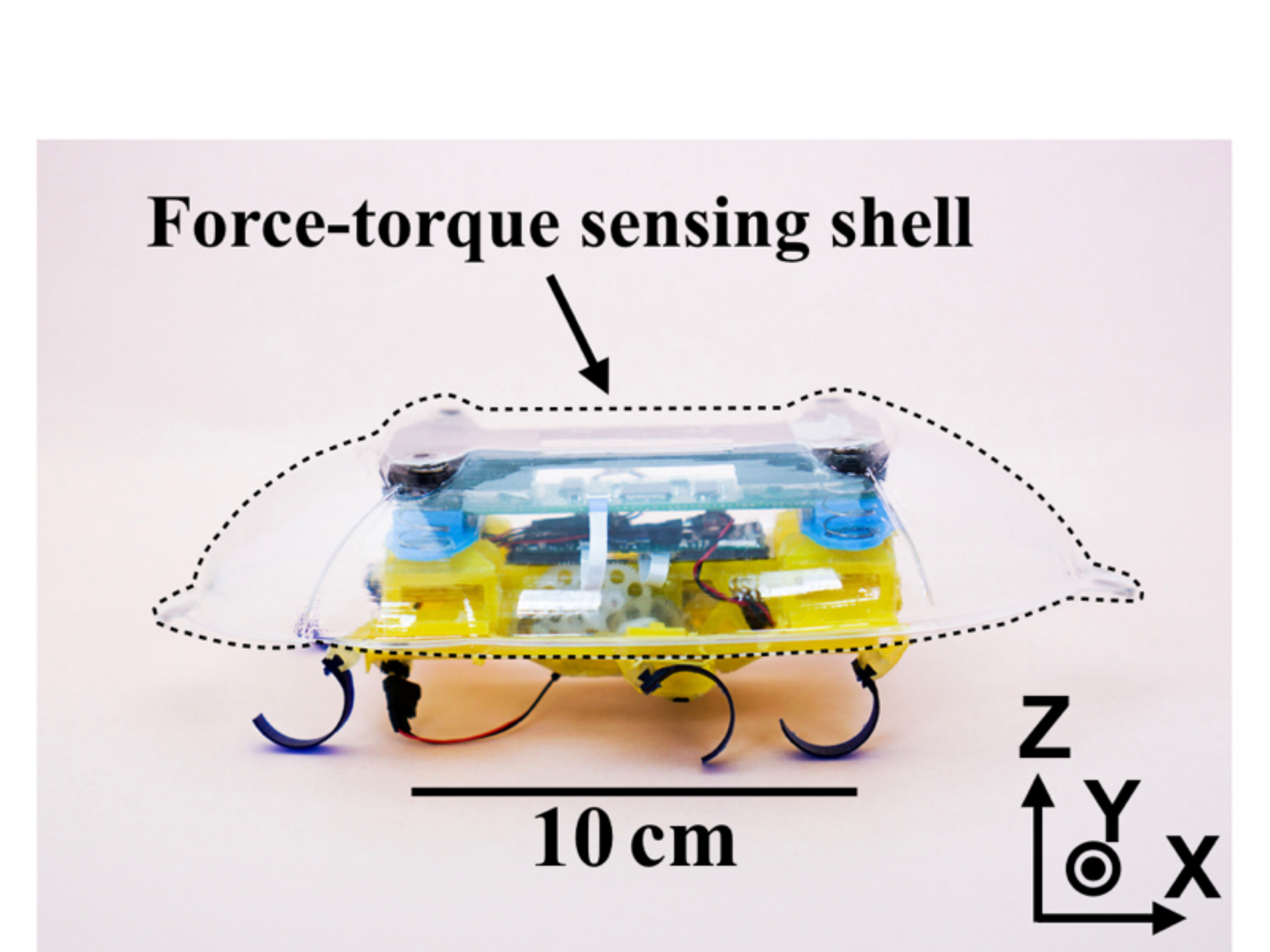} &
        \includegraphics[width=0.47\linewidth]{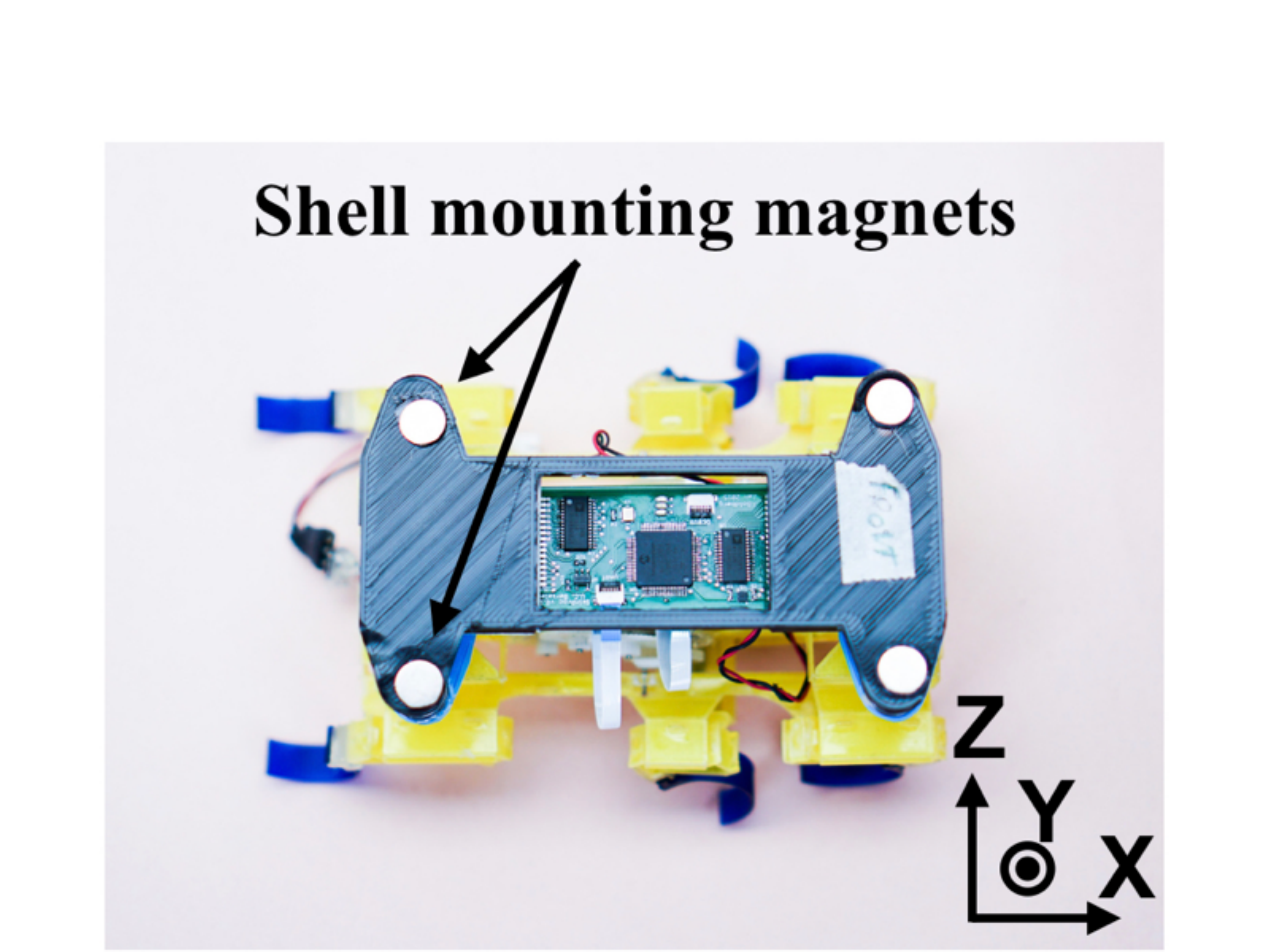}\\
        (a) & (b)\\
    \end{tabular}
\caption{Side (a) and top view (b) of VelociRoACH with force-torque sensing shell.}
\label{fig:side_view_cartoon}
\label{fig:two_sensors}
\end{figure}
\subsection{Robotic platform}
\subsubsection{VelociRoACH}\hfill\\
The base robotic platform, shown in Fig.~\ref{fig:two_sensors}, is the VelociRoACH~\cite{Haldane2013}, a 10~cm hexapedal millirobot developed using the Smart Composite Microstructures (SCM) process as in~\cite{Haldane2015,Hoover2008}. There are 2 computer boards controlling VelociRoACH: ImageProc and SkinProc. ImageProc controls the gait and records telemetry data~\cite{Baek2011}. SkinProc interfaces an optical sensor array that measures the change in displacement between the sensor board and the reflector board to determine the 3D forces and moments acting on the shell~\cite{Goldberg2015}.

\subsubsection{Force-torque sensing shell}\hfill\\
As described in Goldberg et al., on top of the base robotic platform, a force-torque sensing shell (i.e. tactile shell) is attached~\cite{Goldberg2015}. It is composed of three main parts: SkinProc, reflector plate, and shell. The SkinProc board includes a planar array of eight commercially available compact photointerrupters (Sharp GP2S60) with a footprint of 5.44~mm$^2$. Situated on top of the SkinProc board, there are eight reflectors on the 3D printed reflector plate which is attached to the robot's body with springs to allow movement in 3D space. Four reflectors are parallel to the $xy$ plane so that the corresponding photointerrupters are only sensitive to $z$ displacements. Two reflectors are rotated about $y$-axis so they react to $x$- and $z$-axis movements. The remaining two reflectors are rotated about $x$-axis so they react to only $y$- and $z$-axis movements. The reflector plate also contains 4 neodymium magnets that attach to corresponding magnets on the shell. The shell can be detached from the magnets thus allowing access to the SkinProc and ImageProc boards for maintenance. Sensor readings are passed to the ImageProc board at 100~Hz which is then sent to a computer over Xbee wireless radio to be processed off-board.

Goldberg et al. derived a least square solution for predicting forces and torques in $x$-, $y$-, $z$- directions given sensor readings from the eight sensors on the SkinProc board~\cite{Goldberg2015}. In this work, we generated a calibration matrix $C$ from a smaller training data size of 77,522 samples. The root mean squared absolute calibration errors using $C$ on a test data size of 19,932 samples for $F_x$, $F_y$ and $F_z$ are 0.118~N, 0.122~N, and 0.126~N respectively. Zero-input sensor RMS noise for $F_x$, $F_y,$ and $F_z$ are 0.040~N, 0.015~N, and 0.048~N respectively.

\subsection{Specific resistance and drag energy}
We characterize the traversal of the hexapedal millirobot through the cluttered terrain, by its resistance to motion and the efficiency of terrain traversal, measured by drag energy and specific resistance, respectively. 
Drag energy measures the amount of energy the robot exerts when traversing the given terrain and is shown in ~(\ref{eq:drag_energy}).
\begin{equation}
E_{drag} = \overline{F_x}l_{channel}\label{eq:drag_energy}\\
\end{equation}
In~(\ref{eq:drag_energy}), $\overline{F_x}$ is the average drag force resisting the robot's forward motion over the duration of the entire trial and $l_{channel}$ is the length of the channel that the robot traverses.
Specific resistance is a dimensionless quantity that describes the energy efficiency of moving from one point to another. A lower value corresponds with more efficient locomotion. Specific resistance as defined, for example, in Zarrouk et al.~\cite{Zarrouk2013} is shown in~(\ref{eq:specific_resistance}).
\begin{equation}
Specific\;  Resistance\; (\eta) = \dfrac{\overline{P}}{mg\overline{v}}\label{eq:specific_resistance}
\end{equation}
 In~(\ref{eq:specific_resistance}), $\overline{P}$ is the average power consumption of both leg motors, $mg$ is the weight of the robot, and $\overline{v}$ is the average velocity of the robot during the experiment.

\vspace{0.1in}
\section{EXPERIMENT DESIGN}
\vspace{0.1in}
\begin{table}[b]
\squeezeup
\addtolength{\abovecaptionskip}{-3mm}
\caption{Robot and beam obstacle track properties.}
\label{table:platform_properties}
\begin{center}
\begin{tabular}{|c|c|c|}
\hline
 Property & Symbol & Value \\ 
 \hline
 Overall robot length & $l_{robot}$ & 18 cm \\ 
 Overall robot width  & $w_{robot}$ & 11 cm \\
 Overall robot height & $h_{robot}$ & 4.5 cm \\
 Overall robot mass & $m_{robot}$ & 87 g \\
 Channel length & $l_{channel}$ & 28 cm \\
 Beam dimensions & $w\times L\times t$ & $3~cm\times 2.7~cm\times 0.012~cm$ \\
 Friction coefficients & $\mu_s/\mu_k$ & 0.7/0.53 \\
 Flexural modulus & $E$ & 5.3 GPa \\
 Material & - & sheet fiberglass \\
 \hline
\end{tabular}
\end{center}
\end{table}

\begin{table}[b]
\squeezeup
\addtolength{\abovecaptionskip}{-3mm}
\caption{Maximum deflection vs. channel width.}
\label{table:deflection_channel_width}
\begin{center}
\begin{tabular}{|c|c|}
\hline
 Deflection $d$ (cm) & Width $b$ (cm) \\ 
 \hline
 free & free \\ 
 0 & 10 \\
 1 & 8 \\
 2 & 6 \\
 3 & 4 \\ 
 \hline
\end{tabular}
\end{center}
\end{table}

\subsection{Beam obstacle track}
Inspired by the cluttered terrain model from Li et al.~\cite{Chen2015}, we designed and built an obstacle track using compliant beams made out of sheet fiberglass. Geometric and mass properties of the robot and beam obstacle track are summarized in Table~\ref{table:platform_properties}. The compliant fiberglass beams are a laboratory model of thick grass regions found in nature. Thin sheets of fiberglass were cut to form these flaps with dimensions 3~cm$\times$3~cm$\times$0.012~cm with respect to width, length, and thickness. The flaps were punctured 3~mm deep, glued onto two layers of corrugated packaging cardboard, and then attached using neodymium magnets onto blue wax blocks shown in Fig.~\ref{fig:grass_experiment}. The coefficient of static friction was measured from the tangent of the angle at which a weight, covered with fiberglass sheet, started slipping on a tilted surface made out of robot's shell. A similar method was used for measuring the kinetic friction coefficient. The entire channel is 71~cm long and the obstacle track spans 28~cm of the channel length. A total of 11 fiberglass sheets were attached to the cardboard walls and were spaced at 2.5~cm longitudinally along the length of the channel. The beams were placed horizontally to increase surface area of contact with the sides of the shell.

\begin{figure}[t]
\addtolength{\belowcaptionskip}{-5mm}
\centering
\begin{subfigure}[b]{0.48\textwidth}
    \centering
    \includegraphics[width=1\linewidth]{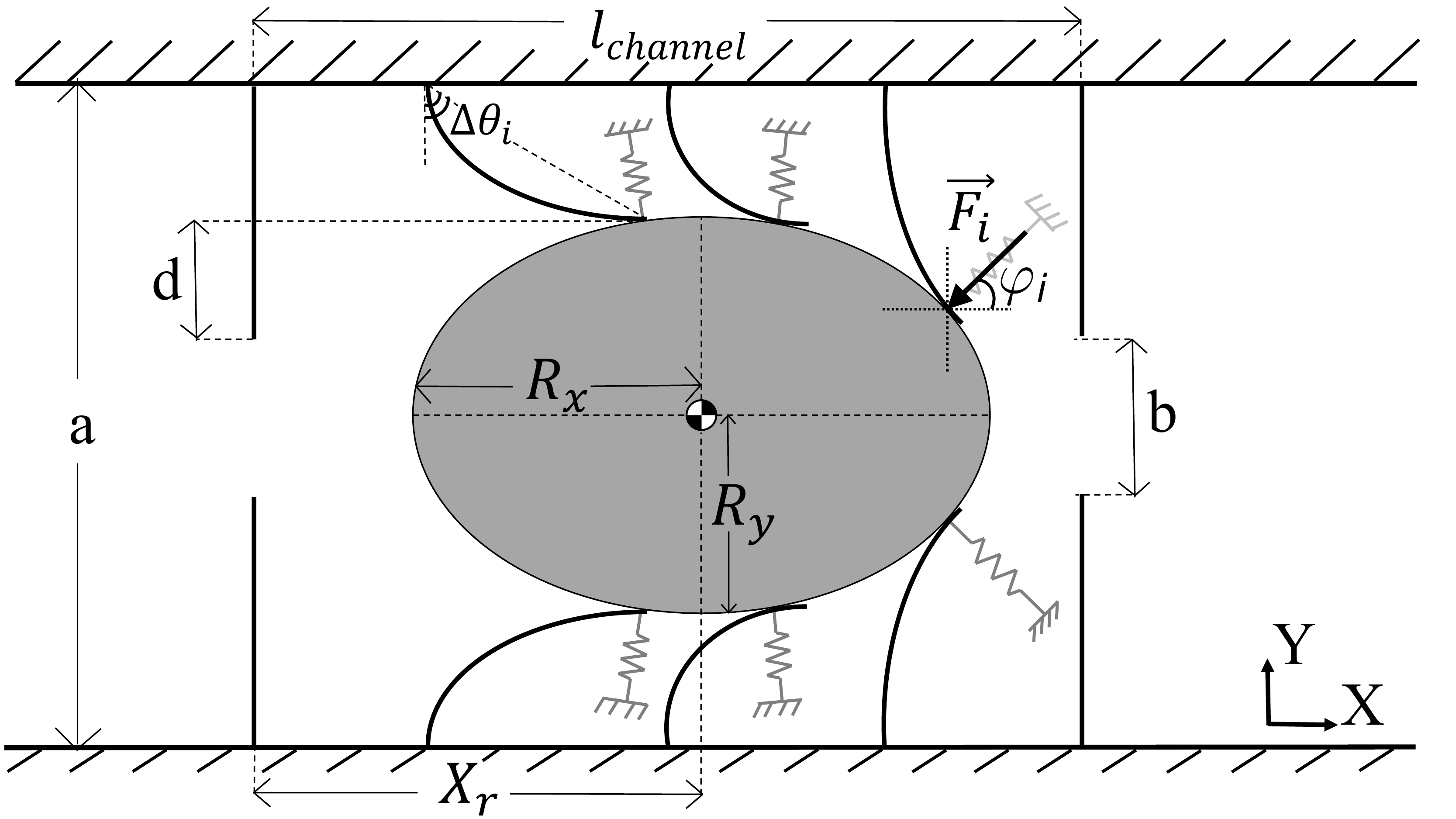}
\end{subfigure}
\caption{Physical model of the cantilevers inside the tunnel deflected due to the shell of the VelociRoACH represented as an ellipse.}
\label{fig:physical_model}
\end{figure}

\subsection{Open-loop trials in variable channel width}
The lateral channel width between opposing beams was varied to create interaction forces of differing magnitudes. Channel width \textit{b} is defined as the distance between the tips of opposite compliant beams in Fig.~\ref{fig:physical_model}. Experimental trials are run for each of 5 possible values of maximum deflection $d$: free, 0 cm, 1 cm, 2 cm, and 3 cm. Maximum deflection is the largest displacement perpendicular to the channel of each beam that contacts the robot. The corresponding channel widths are given in Table~\ref{table:deflection_channel_width}. \par
In experiment trials with the free case, the robot is run outside of the beam obstacle track. In the 0 cm, 1 cm, 2 cm, and 3 cm deflection trials, the robot is placed at the center of the track and run through the channel using an alternating tripod gait with a 1 Hz stride frequency. As the robot does not adjust its gait according to sensory input, this is an open-loop locomotion experiment. During each trial, telemetry data containing leg positions, back EMF signals of the motors, and tactile shell sensor readings are collected. An overhead camera (Nikon D700) records video of each trial. The footage is then used to obtain channel entrance and exit timestamps and also compute average forward velocity $\overline{v}$ of the robot while in the channel. 

\subsection{Quasi-static contact forces}
As shown in the model of Fig.~\ref{fig:physical_model}, the grass-like beams generate a quasi-static force dependent on the deflection caused by the ellipsoidal robot body as it moves through the channel. To obtain a rough bound on contact forces, each grass-like beam is modeled as a torsional spring with linear stiffness $k_t$ as follows:
\begin{equation}
\label{eq:kt_definition}
    k_{t} = \frac{E I}{L},
\end{equation}
where $E$ is the flexural modulus of the beam, $I = \frac{w t^3}{12}$ is the area moment of inertia of the beam for bending about the $z$-axis, and $L$ is the length of the beam. (As can be seen in the accompanying video, the grass-like beams are subject to large out-of-plane deformations with varying contact locations which limit the applicability of this simple spring model.)

Each individual beam, indexed $i \in [1, n]$, produces a force $\mathbf{F_i}$ in the $XY$ plane according to the angular deflection of the beam $\Delta \theta_i$. The angular contact location $\phi_i$ of each beam tip with the ellipsoidal robot body is computed from the furthest forward intersection point of a circle at the base of the beam of radius $L$ with the ellipse. Measured from the first beam, the horizontal contact location $X_i$ of each beam is used to compute $\Delta \theta_i$:
\begin{equation}
\label{eq:hor_beam_contact}
X_i = R_x \cos \phi_i + X_r\\
\end{equation}
\begin{equation}
\label{eq:ang_beam_deflection}
\Delta \theta_i = \sin^{-1}\left(\frac{X_i - l_i}{L}\right),
\end{equation}
where $R_x$ is the major axis radius of the ellipse, $X_r$ is the horizontal center of mass position of the ellipse, and $l_i = \frac{l_{channel}}{n} (i-1)$ is the base position of the beam along the channel.

Now, the ellipse geometry at the contact location can be used to compute unit normal vector $\mathbf{\hat{n}_i}$ pointed into the ellipse and unit tangent vector $\mathbf{\hat{t}_i}$ directed backwards along the channel:
\begin{equation}
\label{eq:unitnormal}
\mathbf{\hat{n}_i} = \frac{-R_y \cos \phi_i \mathbf{\hat{x}} - R_x \sin \phi_i \mathbf{\hat{y}}}{\sqrt{R_x^2 \cos^2 \phi_i + R_y^2 \sin^2 \phi_i}}\\
\end{equation}
\vspace{0.1in}
\begin{equation}
\label{eq:unittangent}
\mathbf{\hat{t}_i} = \frac{-R_x \sin \phi_i \mathbf{\hat{x}} + R_y \cos \phi_i \mathbf{\hat{y}}}{\sqrt{R_x^2 \cos^2 \phi_i + R_y^2 \sin^2 \phi_i}},
\end{equation}
where $R_y$ is the minor axis radius of the ellipse. Finally, the force-deflection relation of each beam modeled as a torsional spring can be applied, with each contact force directed at the edge of the friction cone centered along $\mathbf{\hat{n}_i}$ satisfying the sliding coefficient of friction $\mu_k$. Summing the force of the top $n$ beams along the channel, with a factor of 2 multiplying the result to account for the symmetric bottom beams results in the net drag force $F_{drag}$: 
\begin{gather}
\label{eq:F_beam}
\mathbf{F_i} = \frac{k_t}{L} \Delta \theta_i \left(\mathbf{\hat{n_i}} + \mu_k \mathbf{\hat{t_i}} \right)\\
\label{eq:F_drag}
F_{drag}= 2\sum_{i=1}^{n} \mathbf{F_i} \cdot \mathbf{\hat{x}}.
\end{gather}
\section{EXPERIMENTAL RESULTS}
\begin{figure}[b]
\addtolength{\belowcaptionskip}{-5mm}
\centering
\begin{subfigure}[b]{0.48\textwidth}
    \centering
    \includegraphics[width=0.7\linewidth]{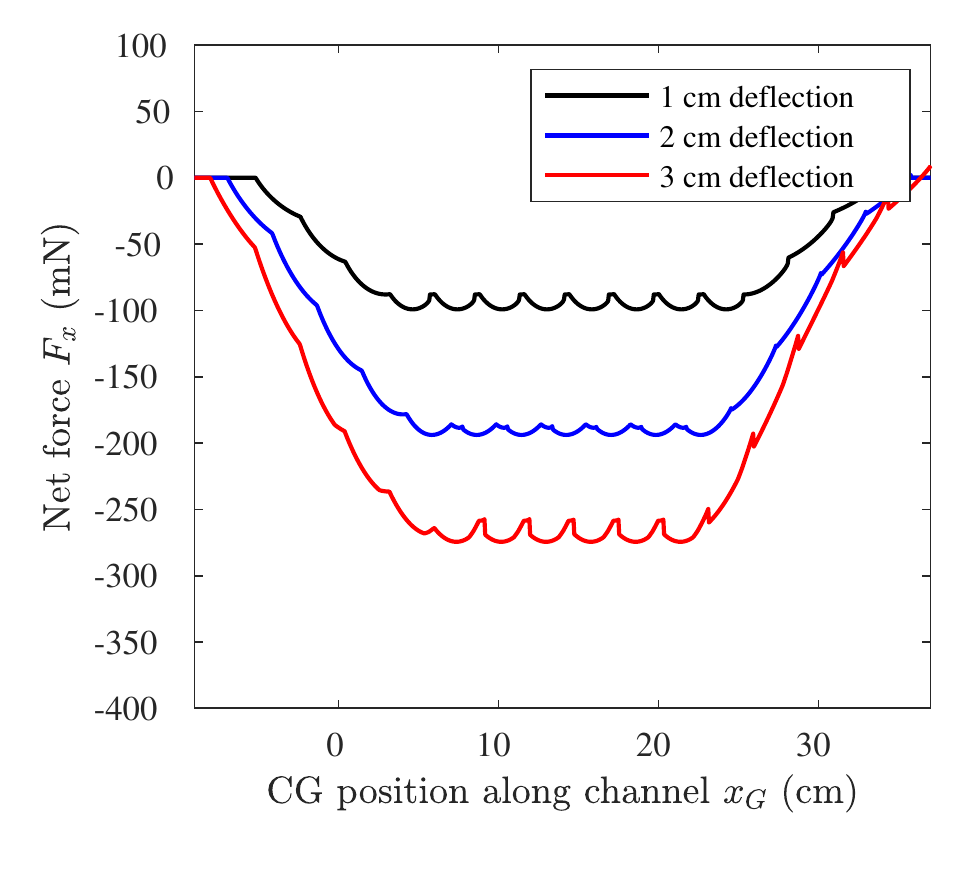}
\end{subfigure}
\caption{Simulated drag forces from quasi-static model for friction $\mu_k = 0.53$ as the robot traverses a channel for 1~cm, 2~cm and 3~cm deflections.}
\label{fig:simulation_drag}
\end{figure}

\subsection{Numerical simulation}
A dynamic numerical simulation based on the quasi-static force analysis using the parameters from the beam obstacle track and the robot was developed. In the simulation, the robot is modeled as an ellipsoid passing through the beams, represented as torsional springs, for different channel widths. Fig.~\ref{fig:simulation_drag} captures the increase in drag forces as the number of beams in contact with the ellipsoid increases. When the maximum number of beams in contact with the ellipsoid is reached, the average drag force no longer increases. \par
Lift forces are not captured in the simulation as only the movement in the $XY$~plane is modeled. When the robot is fully inside the channel, the total number of beams in contact with the robot alternates between 5 and 6 which causes the drag force to oscillate.

\begin{figure}[t]
\addtolength{\belowcaptionskip}{-7mm}
    \begin{subfigure}[b]{0.24\textwidth}
        \centering
        \includegraphics[width=\linewidth]{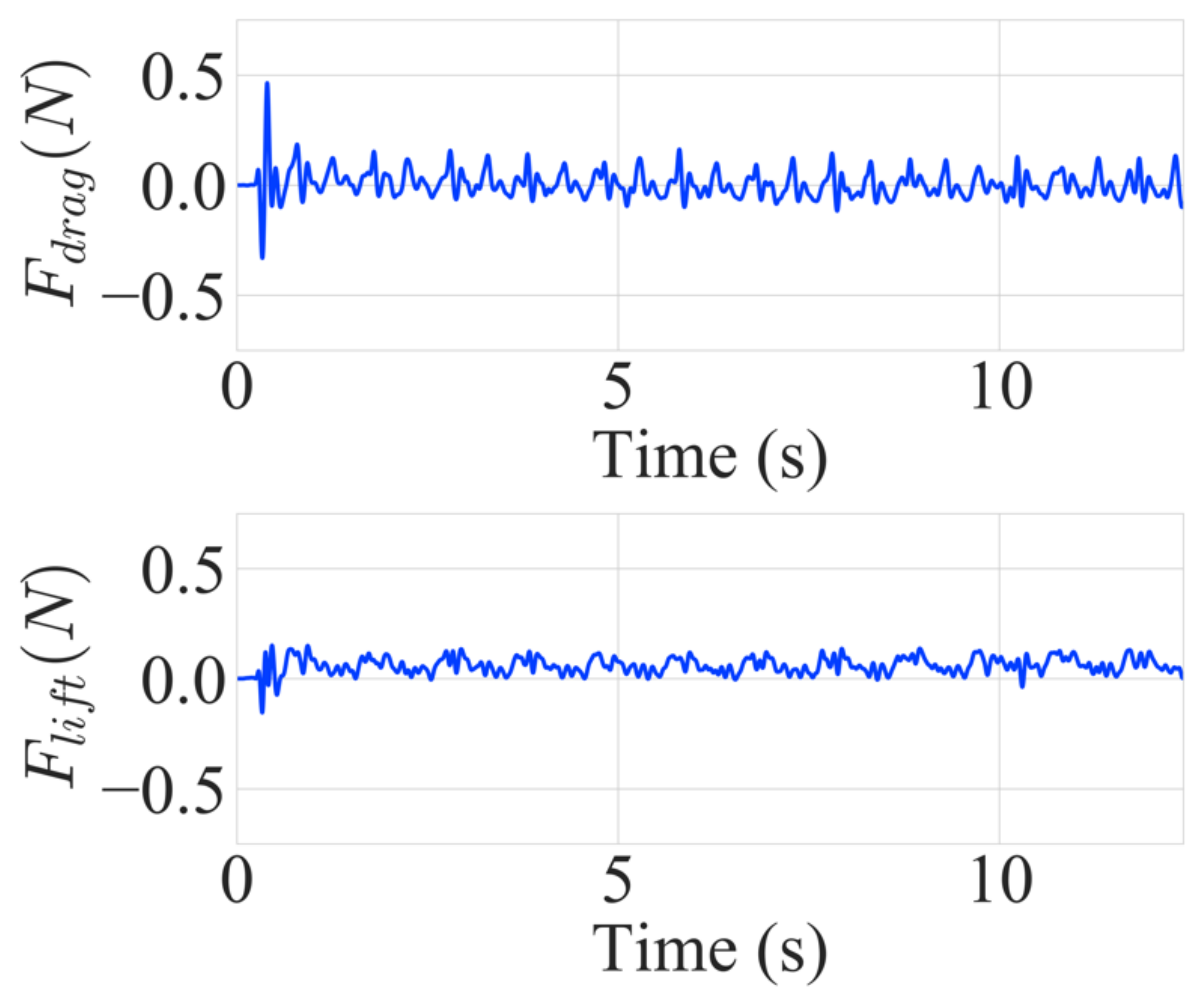}\\
        (a)
    \end{subfigure}%
    \begin{subfigure}[b]{0.24\textwidth}
        \centering
        \includegraphics[width=\linewidth]{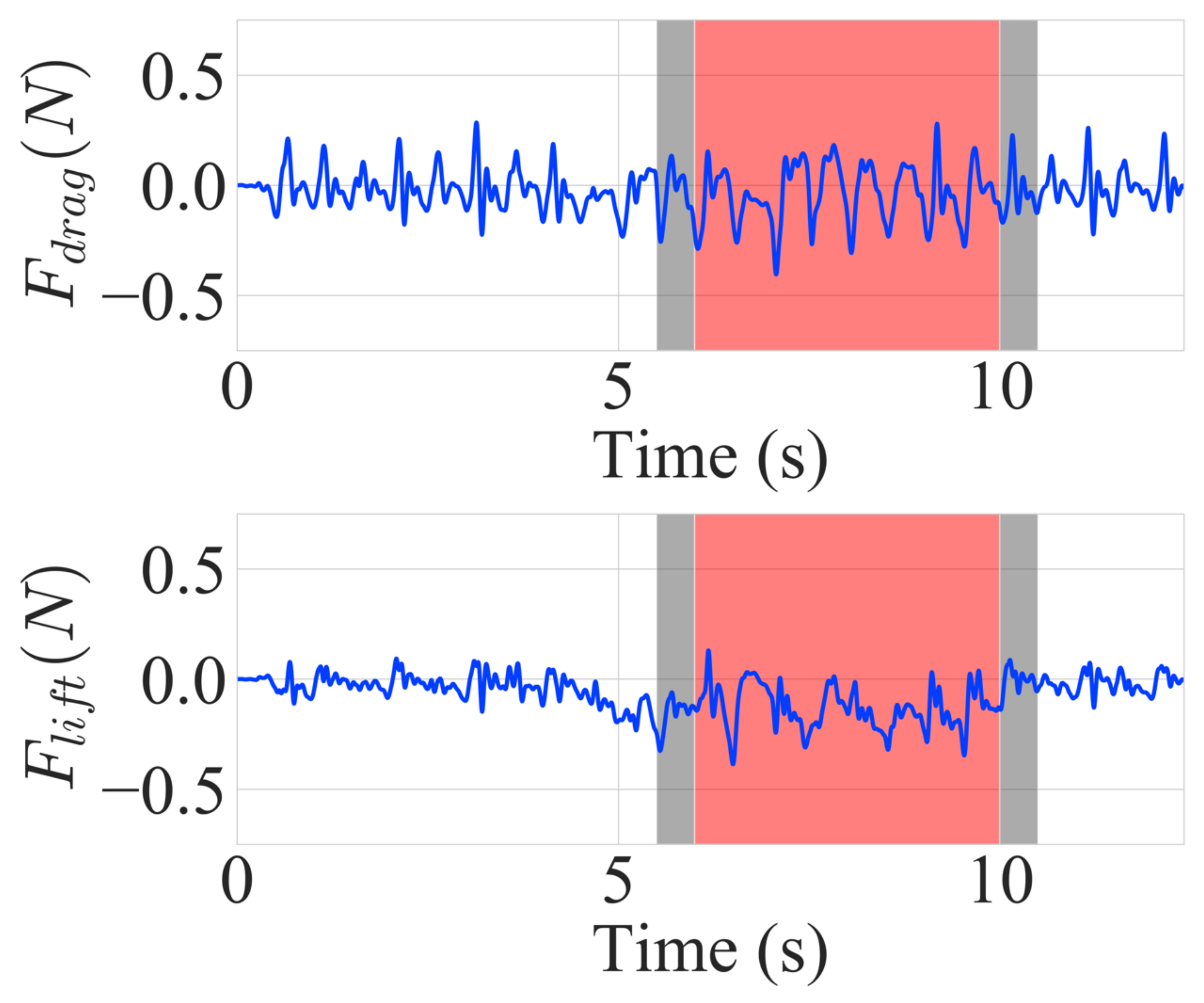}\\
        (b)
    \end{subfigure}\\[1ex]%
    \begin{subfigure}[b]{0.24\textwidth}
        \centering
        \includegraphics[width=\linewidth]{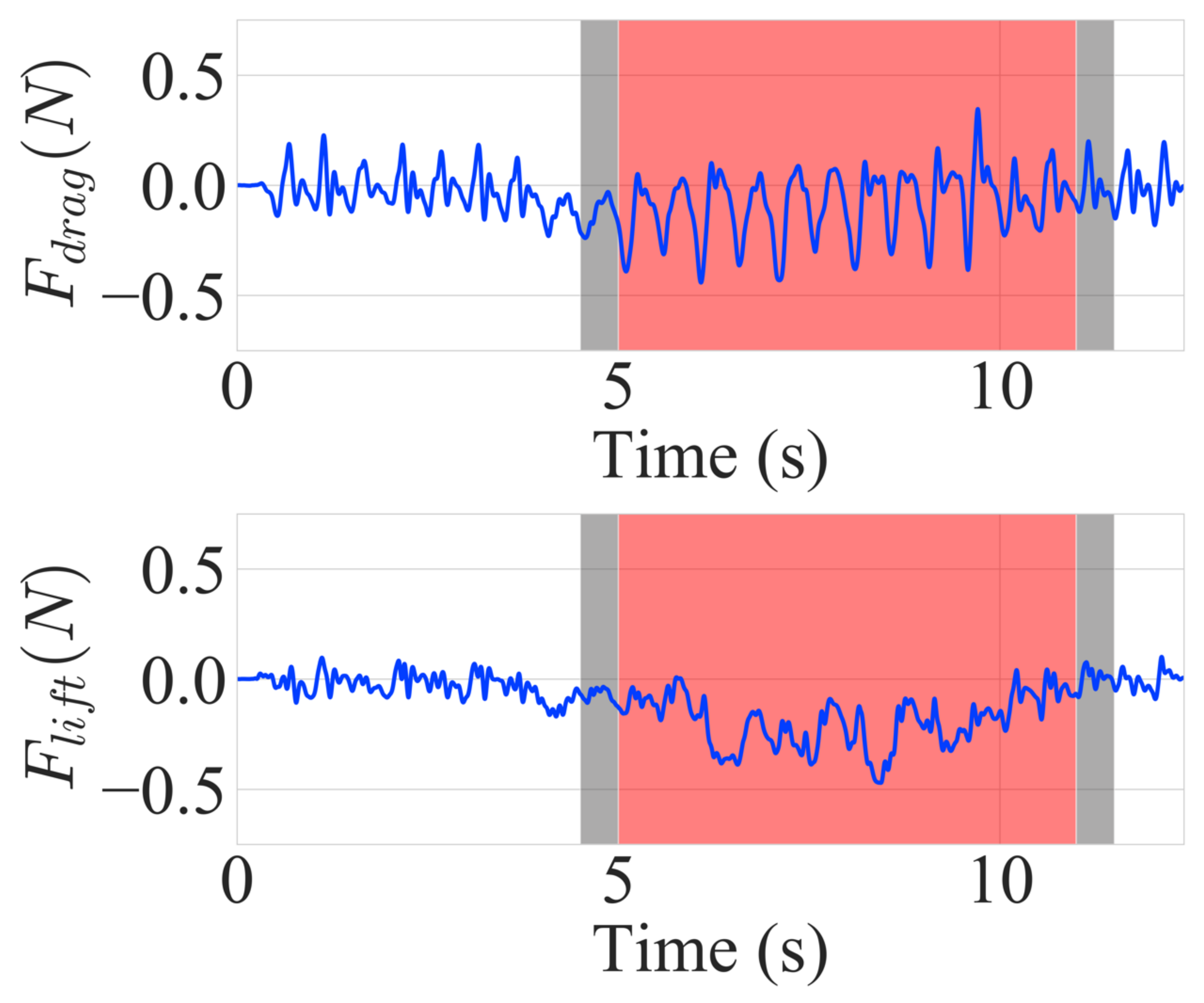}\\
        (c)
    \end{subfigure}%
    \begin{subfigure}[b]{0.24\textwidth}
        \centering
        \includegraphics[width=\linewidth]{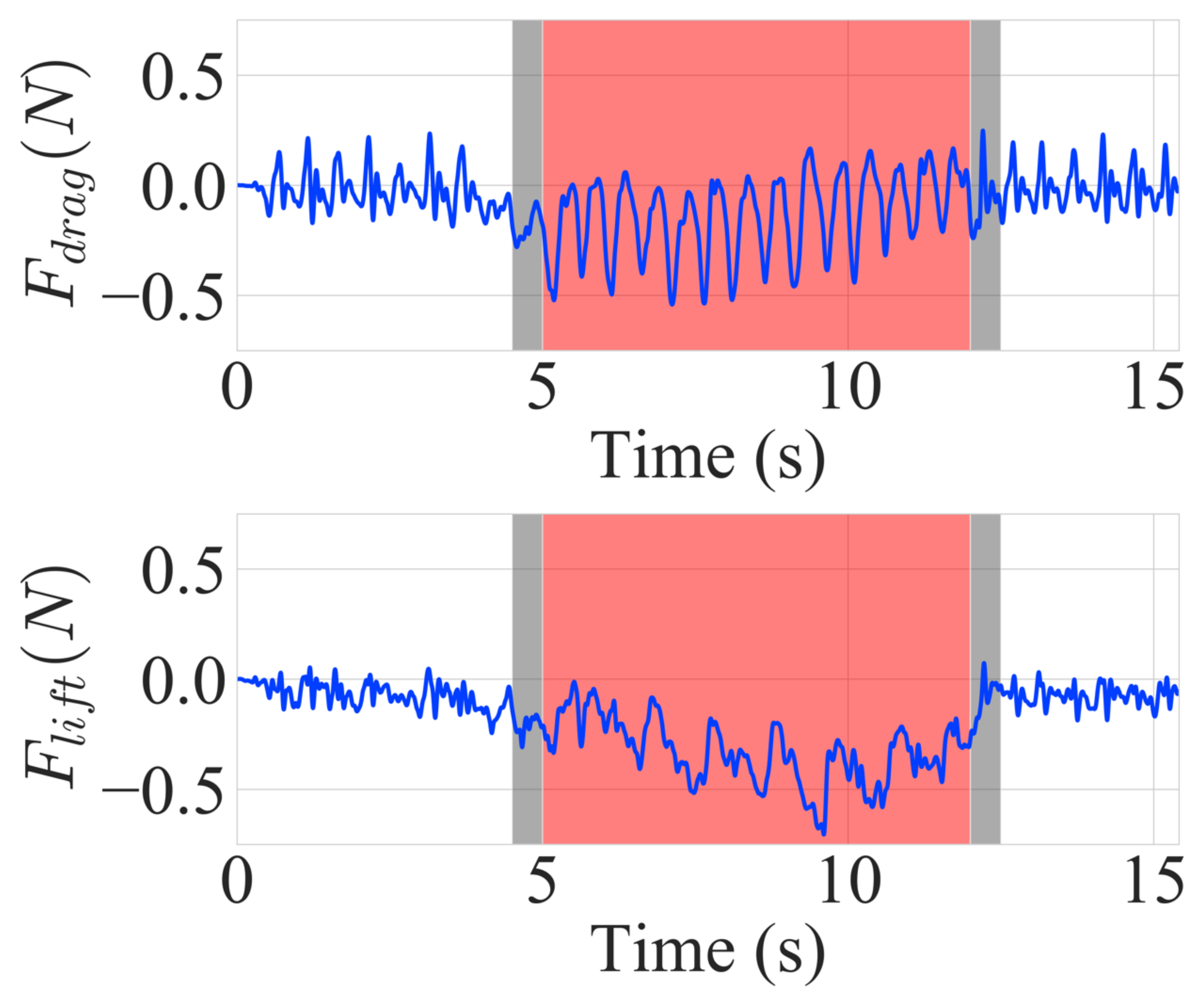}\\
        (d)
    \end{subfigure}%
    \caption{Measured drag and lift forces for deflections: (a) free, (b) 1, (c) 2, (d) 3~cm (0~cm is omitted). Grey regions mark when the robot is entering and exiting the channel. Pink regions mark when the robot is entirely within the channel.}
    \label{fig:drag_lift_plots}
\end{figure}

\subsection{Drag and Lift Force Measurements}
Fig.~\ref{fig:drag_lift_plots} shows the measured drag and lift forces on the tactile shell as the VelociRoACH traverses the row of compliant beams. Like the simulation results in Fig.~\ref{fig:simulation_drag}, Fig.~\ref{fig:drag_lift_plots} shows that average drag forces increase as channel width decreases. However, the simulated average drag forces for 1~cm, 2~cm, and 3~cm deflections overpredict the empirical results of 0.13~N, 0.08~N, 0.03~N respectively. The difference could be explained by a number of factors, including oscillations from leg motions, shell deformations, and negative lift from the contact between the beams and the shell. \par
\begin{figure}[t]
\addtolength{\belowcaptionskip}{-5mm}
\centering
\begin{tabular}{cccc}
    \centering
    \includegraphics[width=.20\linewidth]{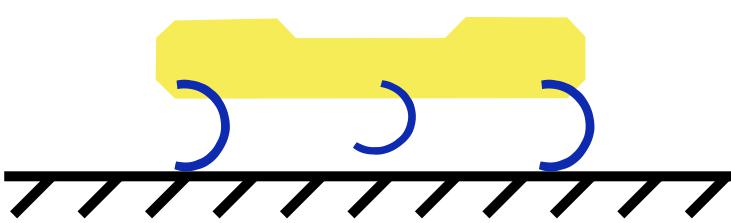} &
    \includegraphics[width=.20\linewidth]{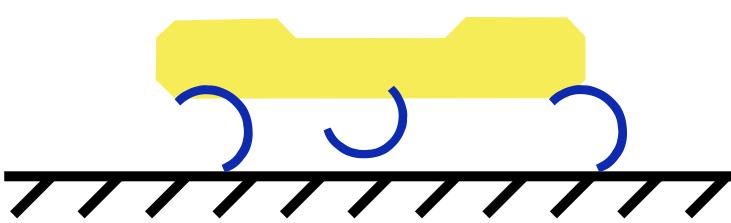} &
    \includegraphics[width=.20\linewidth]{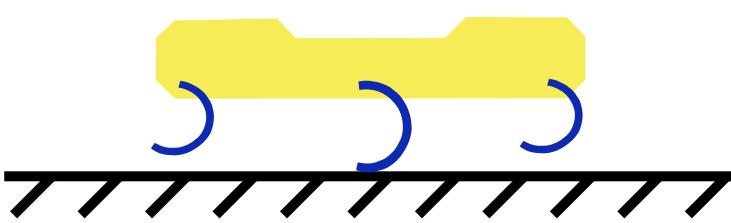} &
    \includegraphics[width=.20\linewidth]{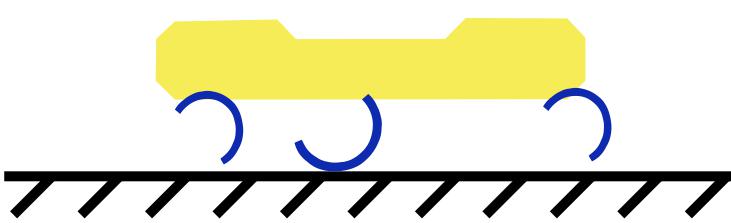}\\
    $0$ & $\frac{\pi}{2}$ & $\pi$ & $\frac{3 \pi}{2}$\\
\end{tabular}
\begin{tabular}{cc}
    \centering
    \includegraphics[width=.45\linewidth]{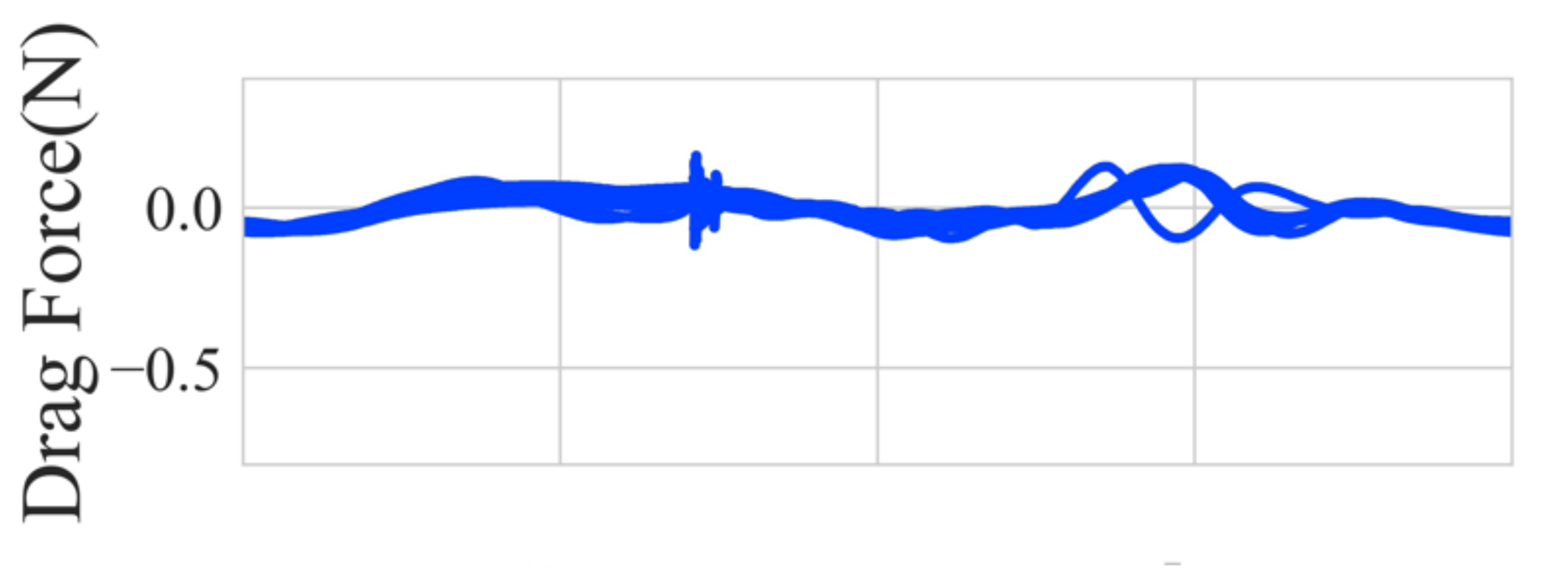} &
    \includegraphics[width=.45\linewidth]{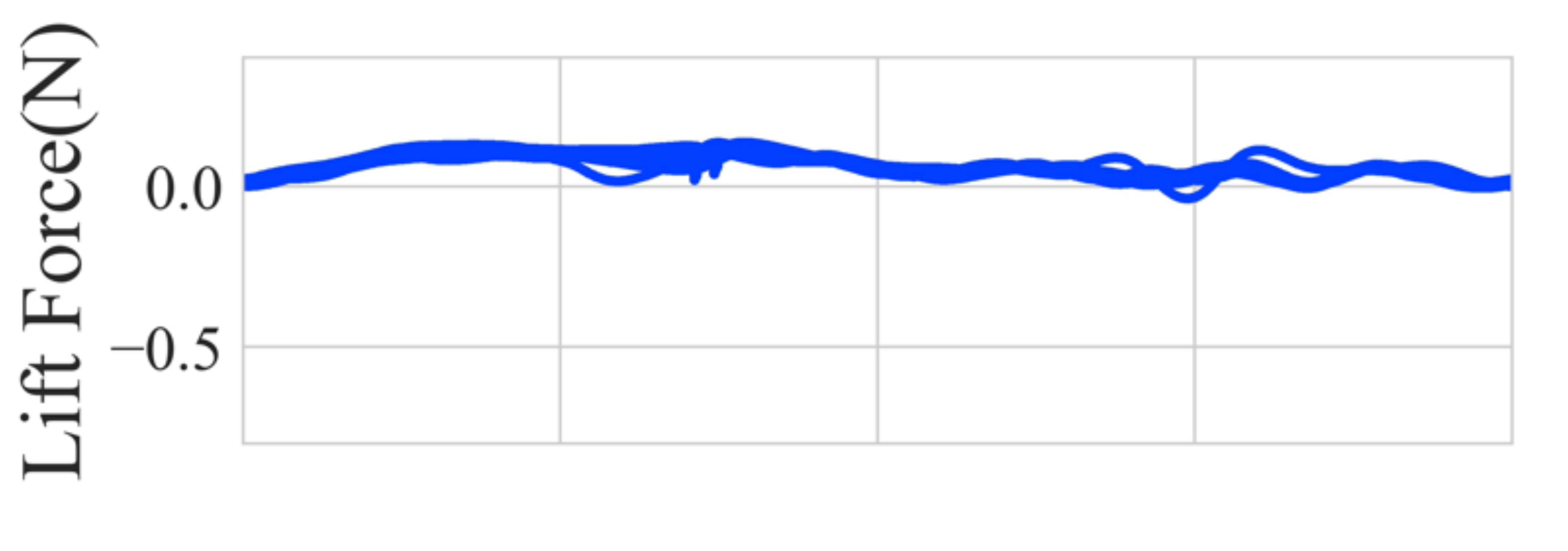}\\
    \includegraphics[width=.45\linewidth]{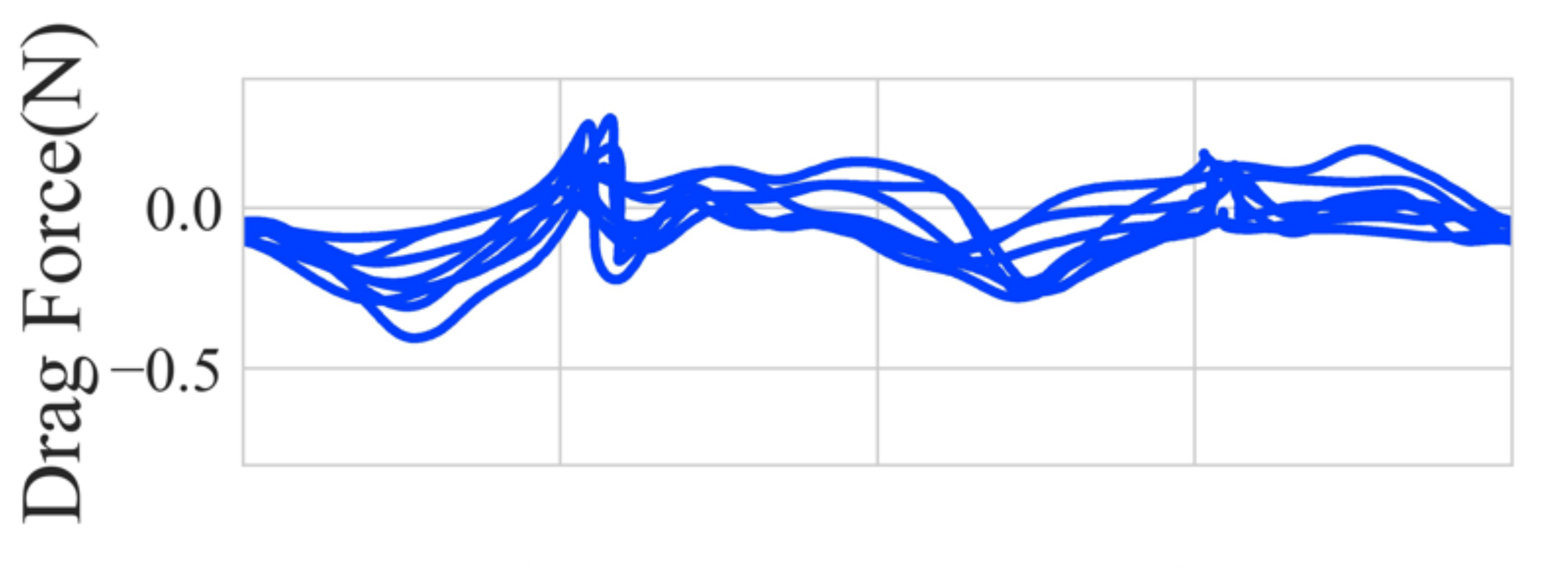} &
    \includegraphics[width=.45\linewidth]{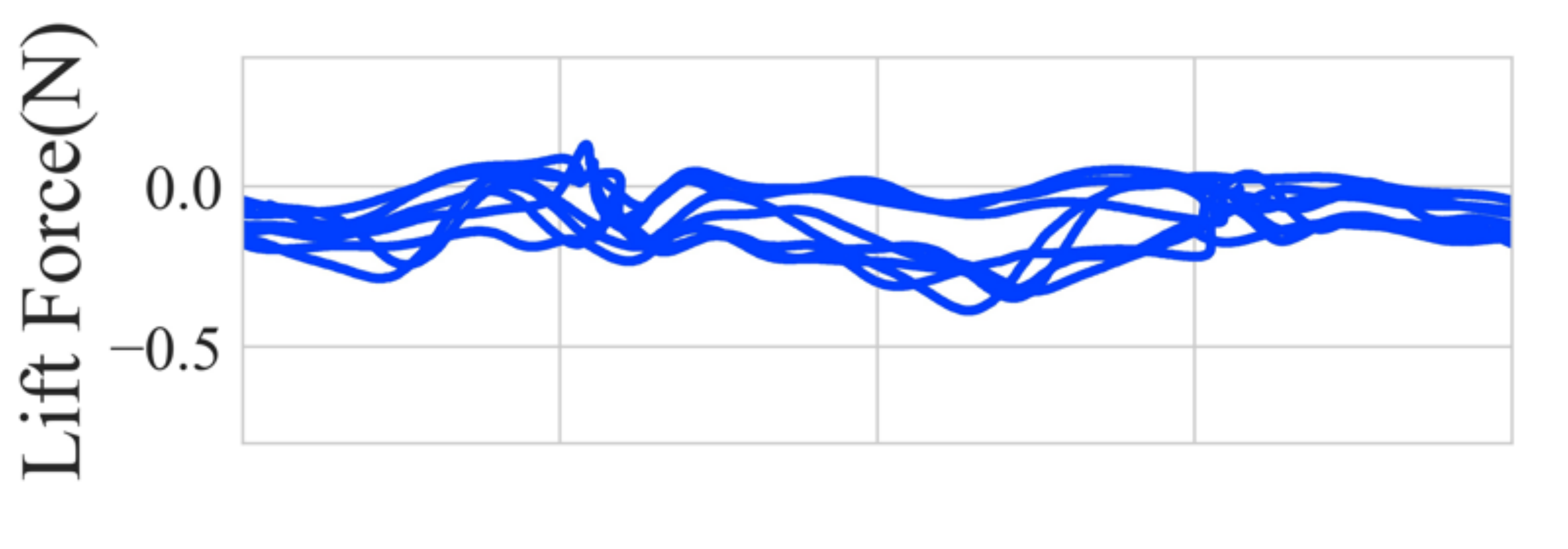}\\
    \includegraphics[width=.45\linewidth]{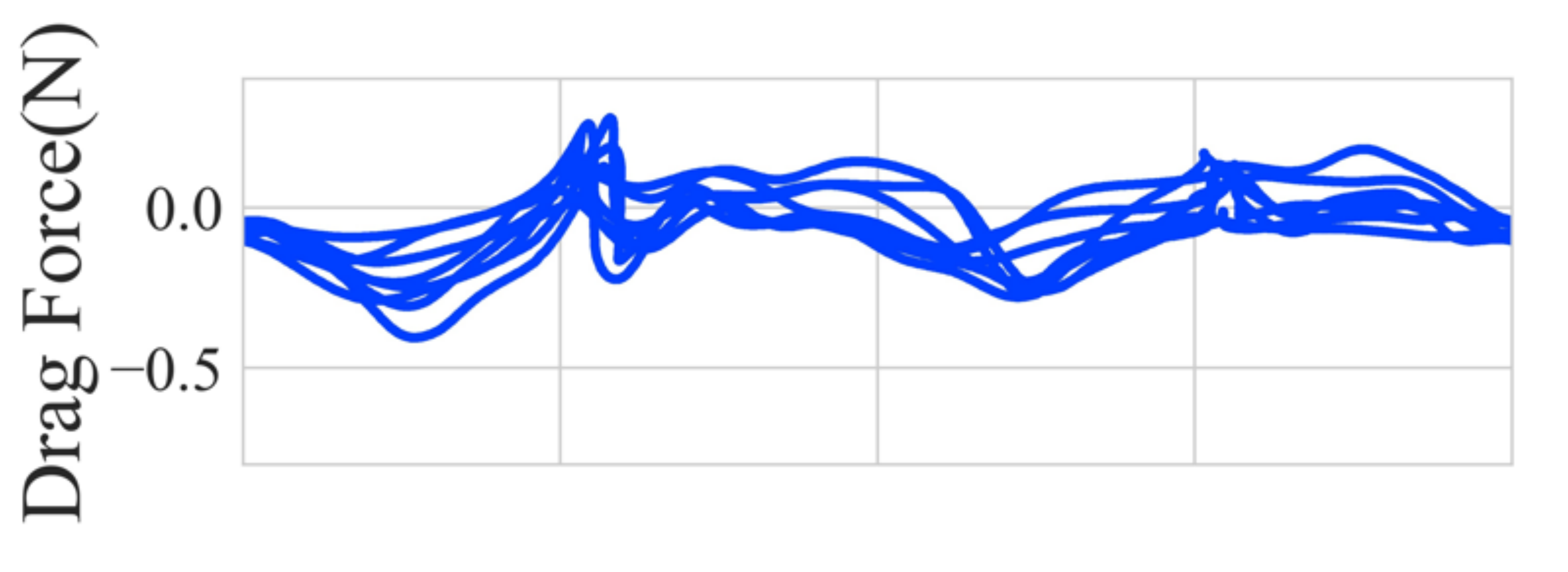} &
    \includegraphics[width=.45\linewidth]{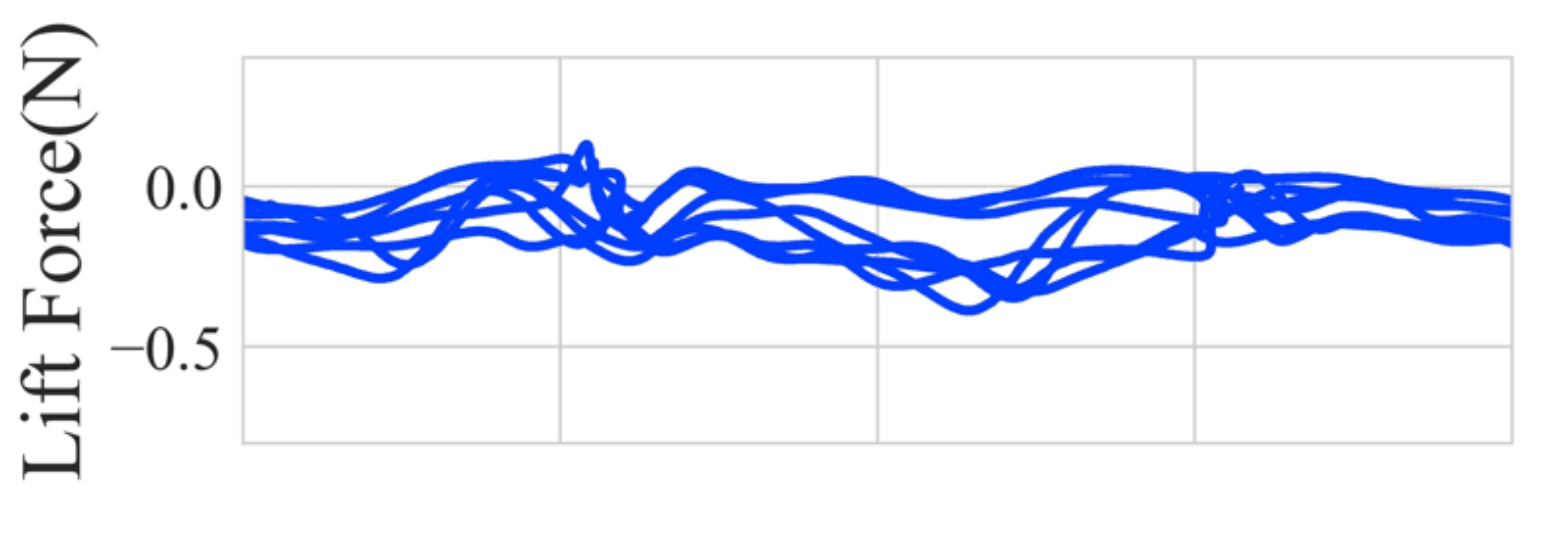}\\
    \includegraphics[width=.45\linewidth]{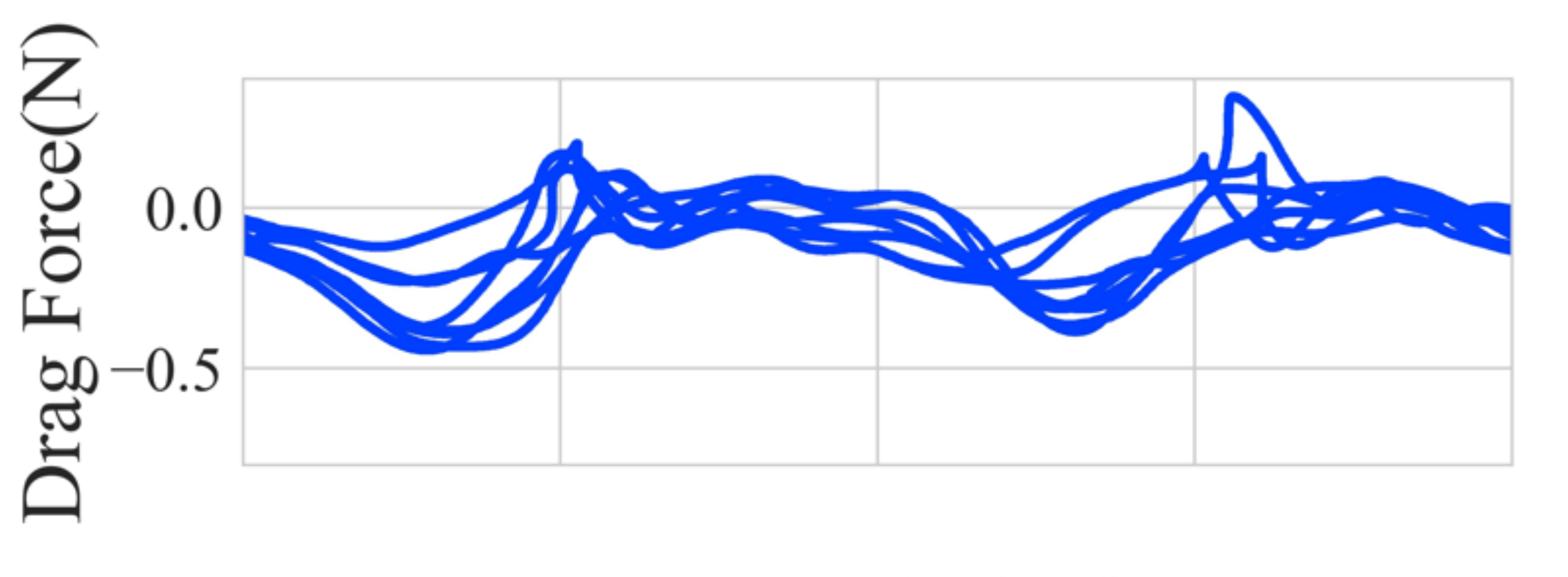} &
    \includegraphics[width=.45\linewidth]{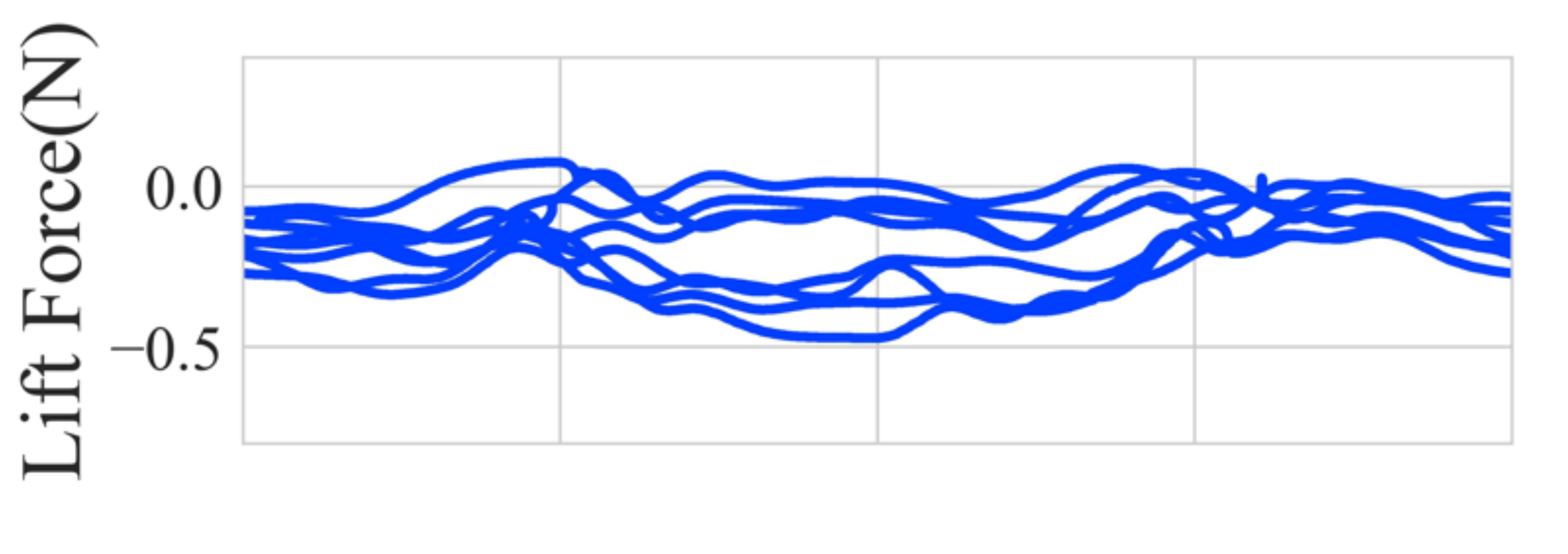}\\
    \includegraphics[width=.45\linewidth]{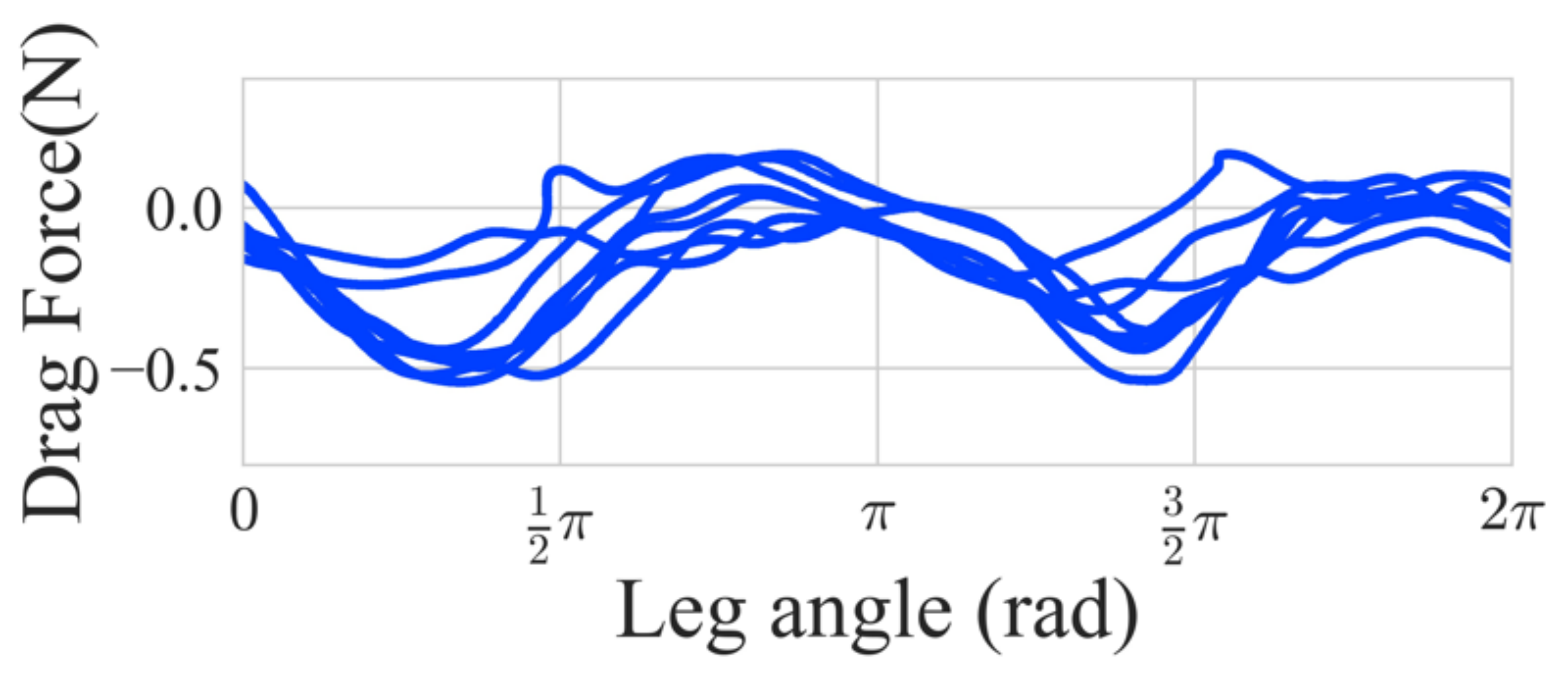} &
    \includegraphics[width=.45\linewidth]{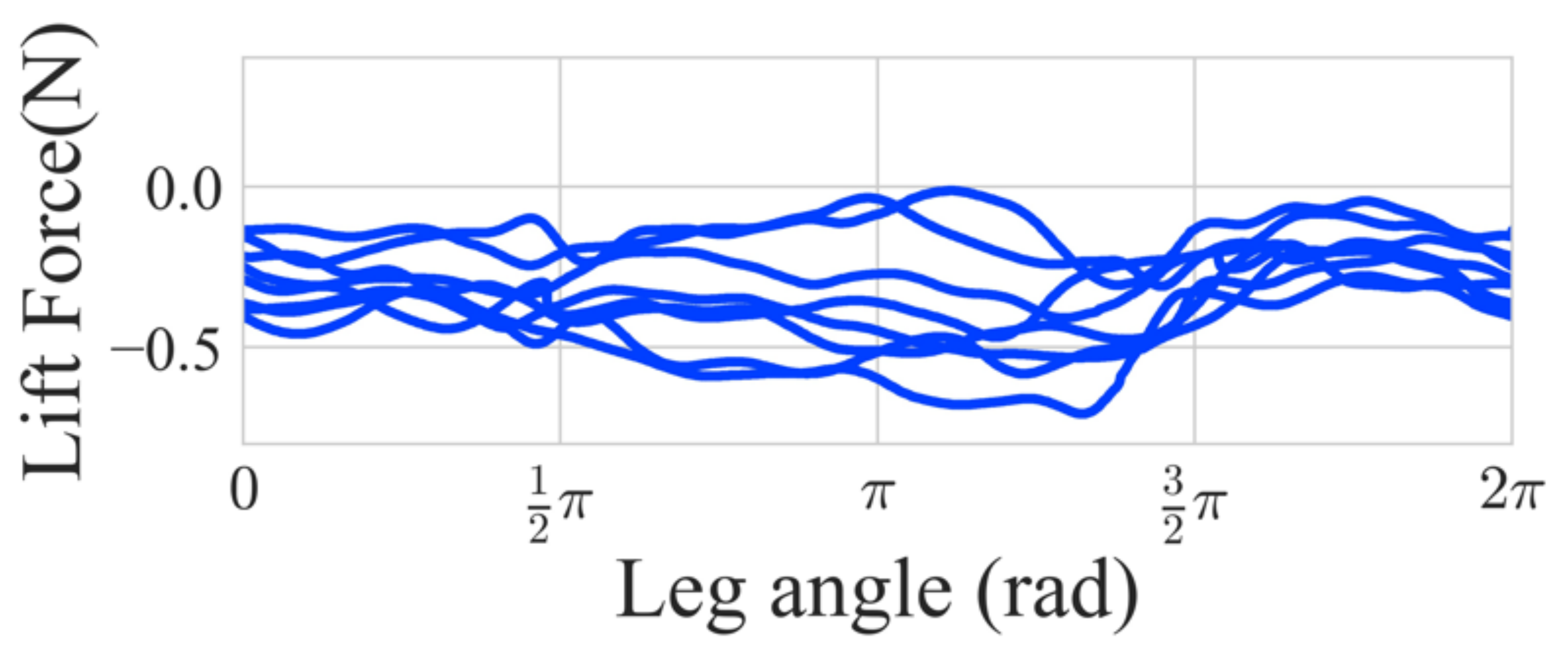}\\
\end{tabular}
\caption{Drag and lift forces plotted against feet positions for all channels taken when VelociRoACH is completely inside the clutter. From top to bottom: free, 0~cm, 1~cm, 2~cm and 3~cm deflections.}
\label{fig:drag_vs_feet_position_plot}
\label{fig:lift_vs_feet_position_plot}
\end{figure}

In Fig.~\ref{fig:drag_lift_plots} and Fig.~\ref{fig:drag_vs_feet_position_plot}, negative lift forces acting on the ellipsoidal shell increase even while the drag forces impeding the motion also increase. This increase in negative lift forces pushing on the shell results in a greater traction in the compliant legs which in turn changes the contact angle $\phi_i$ and unit normal in~(\ref{eq:unitnormal}) of the force as in~(\ref{eq:F_beam}). This could explain the weak coupling between leg positions and lift forces shown in Fig.~\ref{fig:drag_vs_feet_position_plot}. \par
\begin{figure}[t]
\addtolength{\belowcaptionskip}{-5mm}
    \begin{subfigure}{0.48\textwidth}
        \centering
        \includegraphics[width=.9\linewidth]{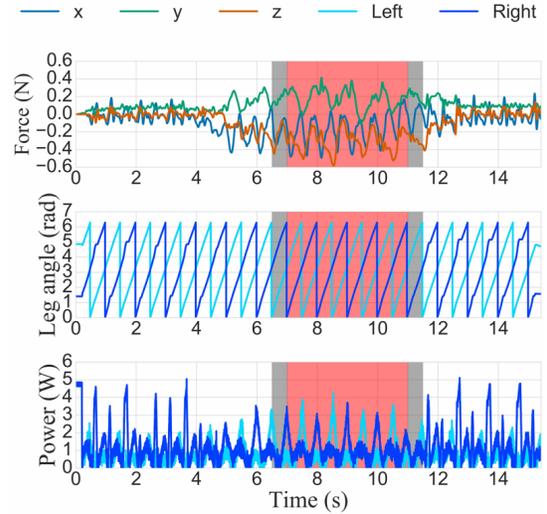}
    \end{subfigure}
    \caption{(Best viewed in color) Telemetry plot from VelociRoACH traversing a channel width of 4~cm (3~cm deflection). From top to bottom, forces in x, y, z directions are measured from tactile sensing shell. Right and left leg positions and power are from robot telemetry.}
    \label{fig:telemetry_tactile_plot}
\end{figure}

The middle plot of Fig.~\ref{fig:telemetry_tactile_plot} and left column plots of Fig.~\ref{fig:drag_vs_feet_position_plot} show that the contact forces on the shell vary with the same period as the leg positions. Forces $F_x, F_y, F_z$, leg positions, and robot power consumption obtained from the telemetry data collected in a 3~cm deflection trial are plotted in Fig.~\ref{fig:telemetry_tactile_plot}. While considerable drag ($-F_x$) and negative lift ($F_z$) forces are observed, the data shows no significant increase in the average $F_y$ force. This is consistent with the distributed contact case matching the physical model in Fig.~\ref{fig:physical_model}, where the lateral forces in $y$-axis are cancelled out. The oscillations in $y$-axis in top plot of Fig.~\ref{fig:telemetry_tactile_plot} and in Fig.~\ref{fig:drag_lift_plots}(a) are most likely due to excitation of the spring-supported shell by leg movements.

\subsection{Energy Cost of Locomotion}
\begin{figure}[t]
\addtolength{\belowcaptionskip}{-5mm}
\centering
 \begin{subfigure}[b]{0.24\textwidth}
 \centering
 \includegraphics[width=\linewidth]{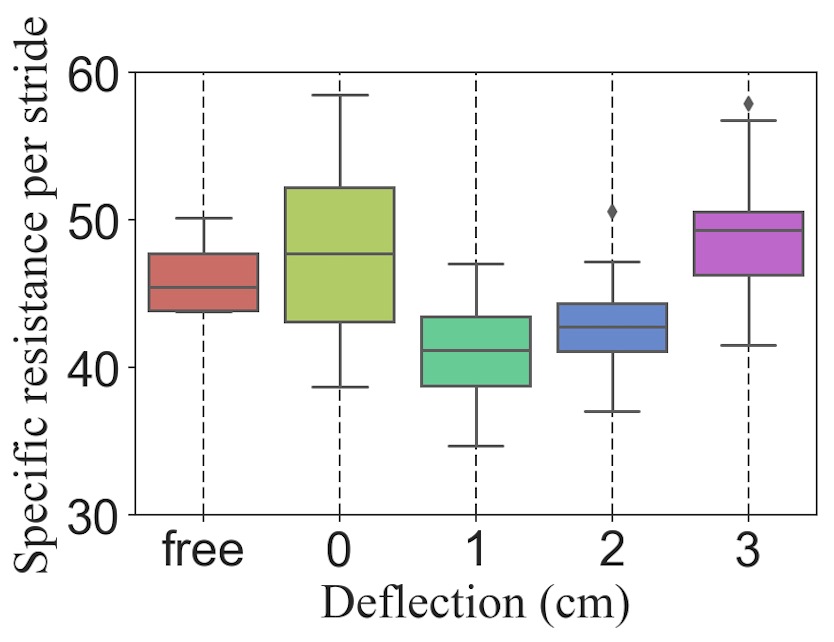}\\
 (a)
 \end{subfigure}\\[1ex]%
 \begin{subfigure}[b]{0.24\textwidth}
 \centering
 \includegraphics[width=\linewidth]{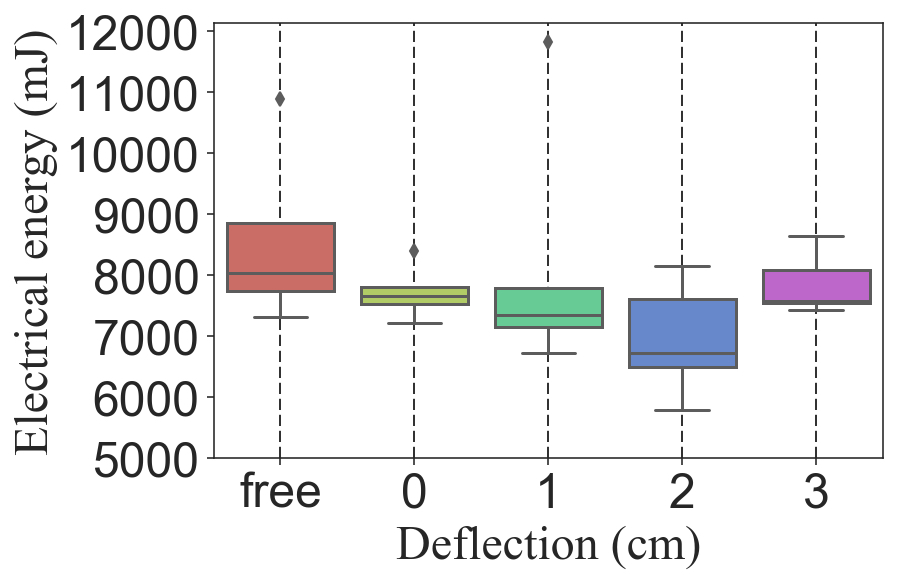}\\
 (b)
 \end{subfigure}%
 \begin{subfigure}[b]{0.24\textwidth}
 \centering
 \includegraphics[width=\linewidth]{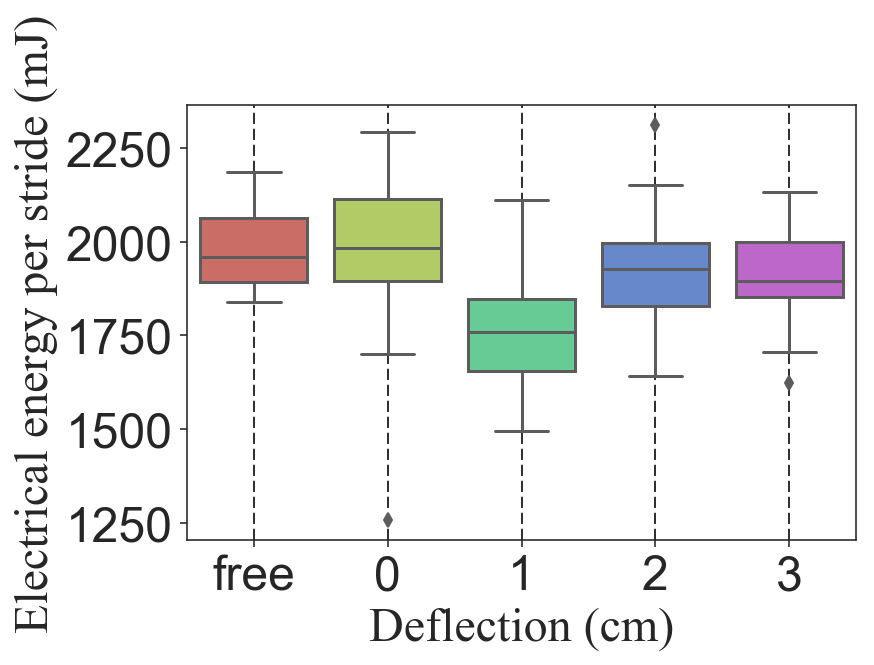}\\
   (c)
 \end{subfigure}\\[1ex]%
 \begin{subfigure}[b]{0.24\textwidth}
 \centering
 \includegraphics[width=\linewidth]{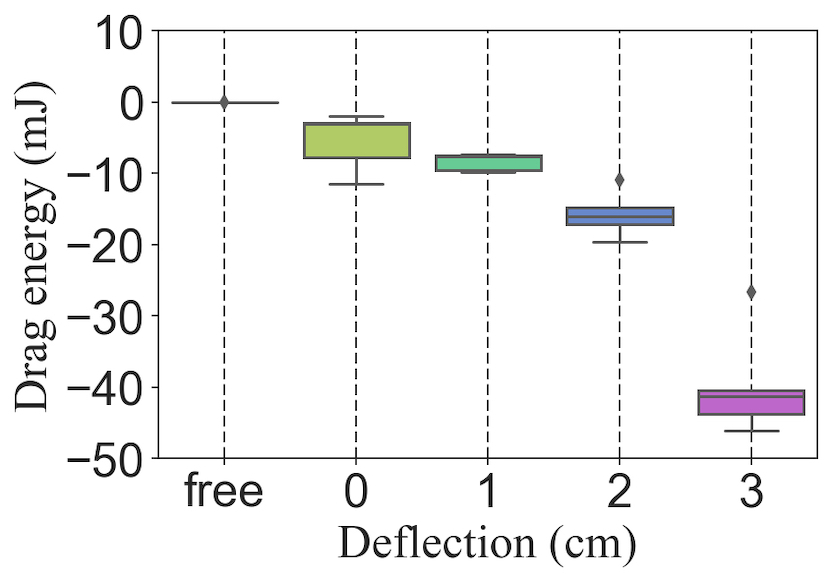}\\
 (d)
 \end{subfigure}%
 \begin{subfigure}[b]{0.24\textwidth}
 \centering
 \includegraphics[width=\linewidth]{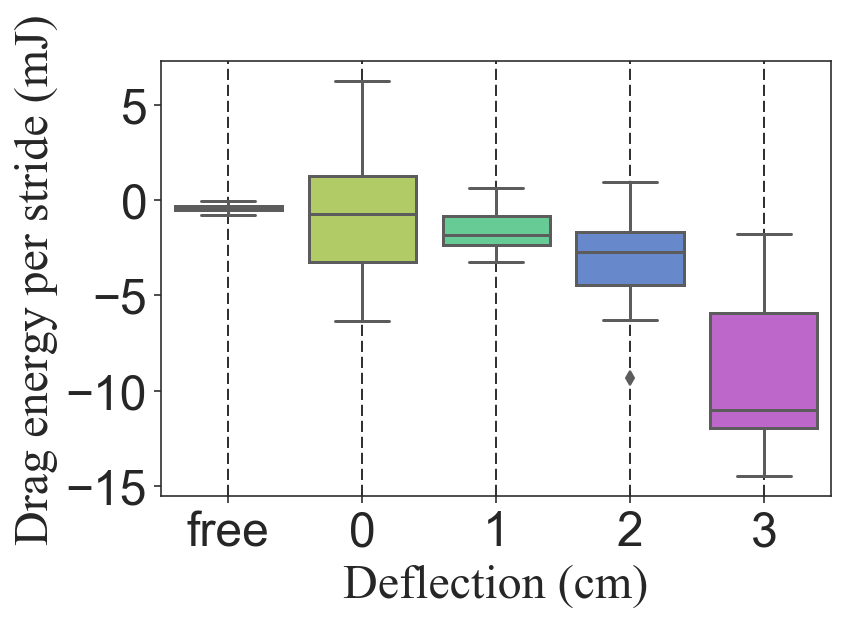}\\
  (e)
 \end{subfigure}%
\caption{Box plots showing drag energy, specific resistance and electrical energy consumed for all trials and varying channel widths.}
\label{fig:top_shell_drag_plot}
\label{fig:top_shell_specific_resistance_plot}
\end{figure}

Using telemetry data from 5 trials for each deflection case, drag energy and electrical energy are plotted per trial and per stride in Fig.~\ref{fig:top_shell_specific_resistance_plot}(b),(c),(d) and (e). Drag energy was computed using~(\ref{eq:drag_energy}) and electrical energy was computed directly from the battery power in telemetry data (bottom plot of Fig.~\ref{fig:telemetry_tactile_plot}). While the electrical energy plots show no perceivable trend with respect to channel deflection, the drag energy plots show a nonlinear increase in the magnitude of drag energy as the channel deflection increases. The plots suggest that the magnitude of drag energy can be used to quantify the densities of terrain clutter, with higher magnitudes corresponding to higher terrain densities.

\subsection{Decrease in resistance to motion}
Fig.~\ref{fig:top_shell_specific_resistance_plot} illustrates that the specific resistance does not monotonically increase as the channel width decreases. The 0 cm deflection specific resistance data show increased spread and an average specific resistance of 48.0, which is closer to the average specific resistance of the 3~cm deflection experiments, 49.5, than to the average specific resistance of 42.9, 41.1, and 45.2 for 2~cm, 1~cm, and \textit{free} experiments. The increased spread in the specific resistance per leg stride for the 0 cm deflection may be explained by the robot’s tendency to veer towards one of the beams and subsequently collide with the wall, reducing forward velocity. \par
Moreover, the fact that the average specific resistance does not strictly decrease with increasing channel width suggests that squeezing through a channel with some downward and backward force can result in less energetic cost than running in the \textit{free} condition. The average specific resistance is lower in the 1cm deflection experiment than in the free experiment. However, if the channel is too narrow, such as in the 3~cm deflection experiment, then running through a channel becomes more energetically costly than running outside a channel. This result suggests that robots may be able to leverage environmental contact to navigate with more energetic efficiency.
\vspace{-5pt}
\section{DISCUSSION AND FUTURE WORK}
This work presents a method to experimentally quantify specific resistance and drag energy of traversing a cluttered terrain. External forces acting on a legged millirobot mounted with a tactile force sensing shell were measured and compared to a model where robot was represented as a rigid ellipsoidal body and the grass-like beams as torsional springs. Each beam exerts a quasi-static force in the $XY$ plane dependent on the deflection caused by the ellipsoidal shell shape. Experimental trials show, while drag and lift forces increased with decreasing channel width, specific resistance did not monotonically increase. This suggests contact between the robot and cluttered terrain could lower specific resistance of locomotion and increase traction compared to free running. \par 
A comparison between electrical energy, specific resistance, and drag energy during the traversal show that reliance on typical telemetry data such as motor power, joint torques, or back EMF is insufficient to distinguish beam deflection conditions. Furthermore, calculating drag energy could provide a better measure in quantifying optimum locomotor paths for cluttered terrains.

Future direction of this work could explore developing a closed loop control of this millirobot by not only utilizing the standard telemetry information such as leg positions and leg power consumption, but also net forces sensed on robot's body. Incorporating force information as an input to learning algorithms could aid millirobots to develop realtime locomotion policies that traverse regions of lower cost of transport in densely cluttered terrains.


\addtolength{\textheight}{-11cm}   



\vspace{-0.1in}
\section*{ACKNOWLEDGMENT}
The authors thank the members of the Biomimetic Millisystems Laboratory at UC Berkeley for their valuable insights and discussions.

\bibliography{shellLift}
\bibliographystyle{ieeetr}

\end{document}